\documentclass{article}

\usepackage{arxiv}

\usepackage[utf8]{inputenc} % allow utf-8 input
\usepackage[T1]{fontenc}    % use 8-bit T1 fonts
\usepackage{hyperref}       % hyperlinks
\usepackage{url}            % simple URL typesetting
\usepackage{booktabs}       % professional-quality tables
\usepackage{amsfonts}       % blackboard math symbols
\usepackage{nicefrac}       % compact symbols for 1/2, etc.
\usepackage{microtype}      % microtypography
\usepackage{lipsum}		% Can be removed after putting your text content
\usepackage{graphicx}
\usepackage[numbers,sort&compress]{natbib}
\usepackage{doi}
\usepackage{adjustbox}
\usepackage{subcaption}

\usepackage{acro}
\usepackage{amsmath,amsfonts,bm}

\def\eqref#1{equation~\ref{#1}}
\def\1{\bm{1}}

\DeclareMathAlphabet{\mathsfit}{\encodingdefault}{\sfdefault}{m}{sl}
\SetMathAlphabet{\mathsfit}{bold}{\encodingdefault}{\sfdefault}{bx}{n}

\DeclareAcronym{st}{
  short = ST ,
  long  = Spatial Transcriptomics
}
\DeclareAcronym{ssl}{
  short=SSL,
  long=Self-Supervised Learning
}
\DeclareAcronym{tsne}{
    short=t-SNE,
    long=t-distributed Stochastic Neighbor Embedding
}
\DeclareAcronym{flops}{
  short=FLOPs,
  long=Floating point operations
}
\DeclareAcronym{mrna}{
  short=mRNA,
  long=messenger RNA
}
\DeclareAcronym{brna}{
  short=bRNA-seq,
  long=Bulk RNA sequencing
}
\DeclareAcronym{scrna}{
  short=scRNA-seq,
  long=Single Cell RNA sequencing
}
\DeclareAcronym{he}{
  short=H\&E,
  long=Hematoxylin and Eosin
}
\DeclareAcronym{sst}{  
    short=SST,
    long=Sequencing-based Spatial Transcriptomics
}
\DeclareAcronym{ist}{
  short=IST,
  long=Imaging-based Spatial Transcriptomics
}
\DeclareAcronym{ffpe}{
  short=FFPE,
  long=Formalin-Fixed Paraffin-Embedded,
}
\DeclareAcronym{tme}{
  short=TME,
  long=Tumor Microenvironment
}
\DeclareAcronym{til}{
  short=TIL,
  long=Tumor-Infiltrating Lymphocyte
}
\DeclareAcronym{tls}{
  short=TLS,
  long=Tertiary Lymphoid Structure
}
\DeclareAcronym{wsi}{
  short=WSI,
  long=Whole-Slide Image
}
\DeclareAcronym{hvg}{
  short=HVG,
  long=Highly Variable Gene
}
\DeclareAcronym{svg}{
  short=SVG,
  long=Spatially Variable Gene
}
\DeclareAcronym{ml}{
    short=ML,
    long=Machine Learning,
}
\DeclareAcronym{sl}{
    short=SL,
    long=Supervised Learning,
}
\DeclareAcronym{dl}{
    short=DL,
    long=Deep Learning,
}
\DeclareAcronym{rl}{
    short=RL,
    long=Representation Learning,
}
\DeclareAcronym{wsl}{
    short=WSL,
    long=Weakly Supervised Learning,
}
\DeclareAcronym{mlp}{
  short=MPL,
  long=Multi-Layer Perceptron
}
\DeclareAcronym{cnn}{
  short=CNN,
  long=Convolutional Neural Networks
}
\DeclareAcronym{vit}{
  short=ViT,
  long=Vision Transformer
}
\DeclareAcronym{mil}{
  short=MIL,
  long=Multiple Instance Learning
}
\DeclareAcronym{msi}{
  short=MSI,
  long=Microsatellite Instability
}
\DeclareAcronym{hrd}{
  short=HRD,
  long=Homologous Recombination Deficiency
}
\DeclareAcronym{tl}{
  short=TL,
  long=Transfer Learning
}
\DeclareAcronym{tcga}{
  short=TCGA,
  long=The Cancer Genome Atlas
}
\DeclareAcronym{gtex}{
  short=GTEx,
  long=Genotype-Tissue Expression
}
\DeclareAcronym{ihc}{
  short=IHC,
  long=Immunohistochemistry
}
\DeclareAcronym{gnn}{
  short=GNN,
  long=Graph Neural Networks
}
\DeclareAcronym{mp}{
  short=MP,
  long=Message Passing
}
\DeclareAcronym{knn}{
  short=k-NN,
  long=k-Nearest Neighbor graph
}
\DeclareAcronym{eg}{
  short=$\epsilon$-graph,
  long=$\epsilon$-Radius graph
}
\DeclareAcronym{gcn}{
  short=GCN,
  long=Graph Convolutional Network
}
\DeclareAcronym{gat}{
  short=GAT,
  long=Graph Attention Network
}
\DeclareAcronym{if}{
  short=IF,
  long=Immunofluorescence
}
\DeclareAcronym{ge}{
  short=GE,
  long=Gene Expression
}
\DeclareAcronym{mg}{
  short=MG,
  long=Marker Gene
}
\DeclareAcronym{pcc}{
  short=PCC,
  long=Pearson Correlation Coefficient
}
\DeclareAcronym{scc}{
  short=SCC,
  long=Spearman Correlation Coefficient
}
\DeclareAcronym{mse}{
  short=MSE,
  long=Mean Squared Error
}
\DeclareAcronym{rmse}{
  short=rMSE,
  long=root Mean Squared Error
}
\DeclareAcronym{r2}{
  short=R²,
  long=R² Score
}
\DeclareAcronym{caf}{
  short=CAF,
  long=Cancer Associated Fibroblats
}
\DeclareAcronym{sccp}{
  short=SCCP,
  long=Single Cell Computational Pathology
}
\DeclareAcronym{fm}{
  short=FM,
  long=Foundation Model
}
\DeclareAcronym{nn}{
  short=NN,
  long=Neural Network
}
\DeclareAcronym{freshf}{
    short=FF,
    long=Fresh Frozen
}
\DeclareAcronym{cp}{
    short=CP,
    long=Computational Pathology
}
\DeclareAcronym{ema}{
    short=EMA,
    long=Exponential Moving Average
}
\DeclareAcronym{kde}{
    short=KDE,
    long=Kernel Density Estimation
}
\DeclareAcronym{cdna}{
    short=cDNA,
    long=complementary DNA
}
\DeclareAcronym{ood}{
    short=OOD,
    long=out-of-domain
}

\title{LEMON: a foundation model for nuclear morphology in Computational Pathology}
\date{}

\author{
  Loïc Chadoutaud\textsuperscript{1,2,3}\thanks{Equal contribution}\\
  \texttt{loic.chadoutaud@curie.fr}
  \And
  Alice Blondel\textsuperscript{1,2,3,*}\\
  \texttt{alice.blondel@minesparis.psl.eu}
  \And
  Hana Feki\textsuperscript{1,2,3}
  \And
  Jacqueline Fontugne\textsuperscript{4,5}\\
  \texttt{jacqueline.fontugne@curie.fr}
  \And
  Emmanuel Barillot\textsuperscript{1,2,3}\\
  \texttt{emmanuel.barillot@curie.fr}
  \And
  Thomas Walter\textsuperscript{1,2,3}\thanks{Corresponding author}\\
  \texttt{thomas.walter@minesparis.psl.eu}
}

\renewcommand{\shorttitle}{LEMON: a foundation model for nuclear morphology in Computational Pathology}

\begin{document}
\maketitle
\vspace{-3em}
\begin{center}
\small
\begin{tabular}{l}
\textsuperscript{1} Institut Curie, F-75005 Paris \\
\textsuperscript{2} Mines Paris PSL, Centre for Computational Biology (CBIO), F-75006 Paris \\
\textsuperscript{3} INSERM, U1331, F-75005 Paris \\
\textsuperscript{4} Institut Curie, U1353/UMR9029 IRIS, Equipe IMPACT, F-75006 Paris \\
\textsuperscript{5} Department of Pathology, Université Paris-Saclay, UVSQ, Institut Curie, Saint-Cloud, France
\end{tabular}
\end{center}
\vspace{3em}

\begin{abstract}
  Representation learning is a central challenge in \ac{cp}, with direct implications for cancer research and precision medicine. While \ac{ssl} has advanced patch and slide-level analysis of \acp{wsi}, single-cell representation learning has remained underexplored, despite its importance for characterizing cell types and phenotypes. We introduce LEMON (Learning Embeddings from Morphology Of Nuclei), a self-supervised foundation model for scalable single-cell image representation. Trained on millions of cell images spanning diverse tissues and cancer types, LEMON provides versatile and robust morphology representations that enable large-scale single-cell studies in pathology. We demonstrate its effectiveness across diverse prediction tasks on five benchmark datasets, establishing LEMON as a new paradigm for cell-level computational pathology. The weights of our model can be accessed at \url{https://huggingface.co/aliceblondel/LEMON}.
\end{abstract}

\section{Introduction}
\label{sec:intro}

The proper function of tissue depends on coordinated interactions of diverse cell types, whose spatial organization is essential for maintaining tissue architecture and whose disruption often signals disease. The composition and interactions of individual cells can influence disease development, and in some cases be predictive of treatment response, providing biomarkers of high clinical value, a prominent example being the \ac{tme}\cite{de_visser_evolving_2023, giraldo_clinical_2019, wang_role_2023}. Computational methods for the analysis of tissues at single-cell resolution are therefore important, both for mechanistic research and for clinical applications.

Recent advances in imaging technologies allow molecular characterization of cells in their spatial context at the transcriptional or protein level.However, these techniques remain costly, and datasets are limited in scale. By contrast, \ac{he} staining - the most widely used modality in pathology - provides rich morphological and spatial information and has generated large-scale image datasets, fueling the development of computational pathology.
 
Self-supervised learning (\ac{ssl}) has been instrumental for computational pathology, as large amounts of unlabeled data are available, while labeled data are often scarce.
Because \acp{wsi} are extremely large (often exceeding 50,000 pixels in width), most \ac{ssl} approaches operate at the patch level, learning embeddings from small tiles cropped from \acp{wsi} \cite{wang_transformer-based_2022, zimmermann_virchow2_2024, saillard_h-optimus-0_2024, chen_towards_2024}. More recent work integrates the aggregation of patch features within the \ac{ssl} framework to produce slide-level representations \cite{lazard_giga-ssl_2022, jaume_multistain_2024, ding_multimodal_2024, vaidya_molecular-driven_2025}. 

To date, however, work on cell-level representation learning remains sparse. Notable examples include DinoBloom, which was designed for hematology cell images from blood smears \cite{koch_dinobloom_2024}, and Volta \cite{nakhli_volta_2024}, a self-supervised contrastive learning model trained on a limited set of tissue-derived cell images. These pioneering works are, however, limited by either their domain specificity or the scale and diversity of their training data.

A prerequisite for training powerful, general-purpose \ac{ssl} models for cells is the assembly of a large-scale, diverse datasets of cell images from \acp{wsi}. Recent advances in nucleus segmentation have made large-scale cell extraction feasible. Models like Hover-Net \cite{graham_hover-net_2019} and CellViT \cite{horst_cellvit_2023, horst_cellvit_2025}, along with highly optimized implementations \cite{liakopoulos_hoverfast_2024}, can now segment millions of cells across various tissue types in a reasonable amount of time. This development opens the door to leveraging state-of-the-art \ac{ssl} paradigms, such as DINOv2 \cite{oquab_dinov2_2024} and MoCov3 \cite{chen_empirical_2021}, which are known to benefit significantly from larger and more diverse training datasets both for natural images \cite{vo_automatic_2024} and for pathology images \cite{chen_revisiting_2025}.

In this work, we capitalize on these advancements to address the gap in cell-level representation learning. We introduce LEMON (Learning Embeddings from Morphology Of Nuclei), a new family of \ac{ssl} models for cell images. Our work not only sets a new state of the art for cell representation learning but also provides a comprehensive blueprint for creating and training such models on this particular type of images. Our main contributions are:
\begin{itemize}
    \item We introduce a family of models trained with several contrastive and non-contrastive \ac{ssl} paradigms to learn powerful representations of nuclear morphology.
    \item We conduct an extensive benchmark evaluation showing that LEMON significantly outperforms prior approaches, including both tile-level and cell-level models, on five distinct downstream tasks, encompassing both cell classification and regression challenges of varying difficulty.
    \item We curate several large-scale cell datasets derived from a diverse cohort of \acp{wsi}, and we systematically study the impact of dataset scale and composition (number of cells, slides, and organs) on downstream performance.
    \item We validate the utility of the learned embeddings by demonstrating improved robustness to staining variability and show that they capture gene-expression patterns as measured by \ac{st}. 
\end{itemize}

\subsection{Self-Supervised Learning}

\noindent\textbf{Contrastive Learning.}
The objective of contrastive learning is to learn an embedding space where different augmented views of the same image (a positive pair) are pulled together, while views from different images (negative pairs) are pushed apart. This is often formalized using the InfoNCE loss \cite{oord_representation_2019}. Given a query view $q$ and a set of key views $\{k_0, k_1, ..., k_N\}$, where $k_0$ is the positive key corresponding to $q$ and $\{k_i\}_{i=1}^N$ are negative keys, the loss is defined as:
\begin{equation}
    \mathcal{L}_{q} = -\log \frac{\exp(\text{sim}(q, k_0) / \tau)}{\sum_{i=0}^{N} \exp(\text{sim}(q, k_i) / \tau)}
\end{equation}
Here, $\text{sim}(\cdot, \cdot)$ is a similarity function (e.g., cosine similarity) and $\tau$ is a temperature hyperparameter. While early methods struggled with strategies for sourcing effective negative pairs \cite{xu_negative_2022}, modern approaches like \textbf{MoCov3} \cite{chen_empirical_2021} have streamlined this process. MoCov3 simplifies its predecessors by using a large batch of in-flight samples for the negative keys, removing the need for a dedicated memory bank. It retains the teacher-student framework, where the student network (which produces the query $q$) is trained via backpropagation, and the teacher network (which produces the key $k$) is an \ac{ema} of the student's weights. This momentum-based update ensures the keys remain consistent, which is crucial for stable training.

\noindent\textbf{Non-Contrastive Learning.}
Non-contrastive methods learn representations by matching the outputs for different views of the same image, thereby avoiding the need for explicit negative pairs. To prevent representational collapse, they employ a teacher-student architecture where the teacher's weights $\theta_t$ are an \ac{ema} of the student's weights $\theta_s$. The state-of-the-art \textbf{DINOv2} \cite{oquab_dinov2_2024} model advances this paradigm by composing three distinct loss functions: an image-level matching loss (from DINO) \cite{caron_emerging_2021}, a patch-level masked-image modeling loss (from iBOT) \cite{zhou_ibot_2022}, and a regularization loss (KoLeo). 

The core objective, inherited from \textbf{DINO} \cite{caron_emerging_2021}, operates on multiple views of an image generated by a multi-crop augmentation strategy. This strategy produces a set of high-resolution "global" views ($\mathcal{V}_g$) and a set of lower-resolution "local" views ($\mathcal{V}_l$). The teacher network processes only the global views, while the student network processes all views. The goal is for the student's output distribution $P_s$ to match the teacher's sharpened and centered distribution $P_t$. The loss is composed of two cross-entropy terms:
\begin{itemize}
    \item \textbf{Global-to-Global Matching:} The student's output for one global view is matched against the teacher's output for the other global view.
    \begin{equation}
    \mathcal{L}_{g \leftrightarrow g} = -\sum_{\substack{v, v' \in \mathcal{V}_g \\ v \neq v'}} P_t(v') \log P_s(v) 
    \end{equation}
    \item \textbf{Local-to-Global Matching:} The student's output for each local view is matched against the teacher's output for all global views.
    \begin{equation}
    \mathcal{L}_{l \leftrightarrow g} = -\sum_{v_l \in \mathcal{V}_l} \sum_{v_g \in \mathcal{V}_g} P_t(v_g) \log P_s(v_l)
    \end{equation}
\end{itemize}
The total image-level loss is $\mathcal{L}_{\text{DINO}} = \mathcal{L}_{g \leftrightarrow g} + \mathcal{L}_{l \leftrightarrow g}$. 

To encourage the learning of fine-grained local features, DINOv2 integrates the masked-image modeling objective from \textbf{iBOT} \cite{zhou_ibot_2022}. For an input image, a random subset of its patches $\mathcal{M}$ is masked. The teacher receives the full, unmasked image, while the student receives the masked image. The student is then tasked with predicting the teacher's output tokens for the masked patches. The loss is a cross-entropy between the student's predictions and the teacher's distribution for the masked patches:
\begin{equation}
\mathcal{L}_{\text{iBOT}} = -\sum_{x_m \in \mathcal{M}} P_t(x_m) \log P_s(x_m)
\end{equation}

To further prevent feature collapse and promote feature space uniformity, DINOv2 adds the \textbf{KoLeo} regularizer. This loss acts on the batch of embeddings produced by the teacher network. It encourages diversity by penalizing similarity between distinct images by maximizing the distance between an embedding $z_i$ and its nearest neighbor $z_{NN(i)}$ in the batch. The loss is formulated as:
\begin{equation}
\mathcal{L}_{\text{KoLeo}} = \sum_{z_i \in \text{batch}}-\log \|z_i - z_{NN(i)}\|_2^2
\end{equation}
This pushes embeddings apart, ensuring the feature space does not collapse to a small volume. The final DINOv2 loss is a weighted sum of these three components: $\mathcal{L} = \mathcal{L}_{\text{DINO}} + \alpha\mathcal{L}_{\text{iBOT}} + \beta\mathcal{L}_{\text{KoLeo}}$. 

\subsection{\ac{ssl} in Histopathology}

The paradigms of self-supervised learning, proven effective on natural images, have been widely adapted for computational pathology, predominantly at the level of image tiles or patches. These methods can be broadly categorized by their core \ac{ssl} strategy and their operational scale.

\noindent\textbf{Tile-Level Representation Learning.}
Early successes in pathology \ac{ssl} often involved adapting contrastive learning frameworks. For instance, \textbf{CTransPath} \cite{wang_transformer-based_2022} modified the MoCo framework by defining positives not only as augmented views of the same tile, but also as other tiles with high cosine similarity, which reflects the idea that tiles are much more similar to each other than natural images. More recently, the field has shifted towards non contrastive methods. Inspired by the success of DINOv2, models like \textbf{UNI} \cite{chen_towards_2024}, \textbf{H-Optimus-0} \cite{saillard_h-optimus-0_2024}, and \textbf{Virchow2} \cite{zimmermann_virchow2_2024} have demonstrated the benefits of pre-training Vision Transformers on massive, curated datasets comprising millions of histopathology tiles. The primary contribution of these works lies in meticulous data curation and demonstrating remarkable scaling laws, where model performance on diverse downstream tasks consistently improves with the scale of the pre-training data.

\noindent\textbf{Cell-Level Representation Learning.}
In contrast to the extensive work at the tile level, research on single-cell level representations remains limited. To our knowledge, \textbf{Volta} \cite{nakhli_volta_2024} is the only \ac{ssl} model specifically designed for this task in the context of histopathology. Volta introduces a novel dual-contrastive learning objective. The first objective is a standard instance-discrimination task, where augmented views of the same cell are pulled together in the embedding space. The second objective contrasts the cell's representation against a representation of its local tissue microenvironment, which is obtained from the surrounding background patch after masking cell nuclei.

\section{Methods}

\begin{figure}
    \centering
    \includegraphics[width=\linewidth]{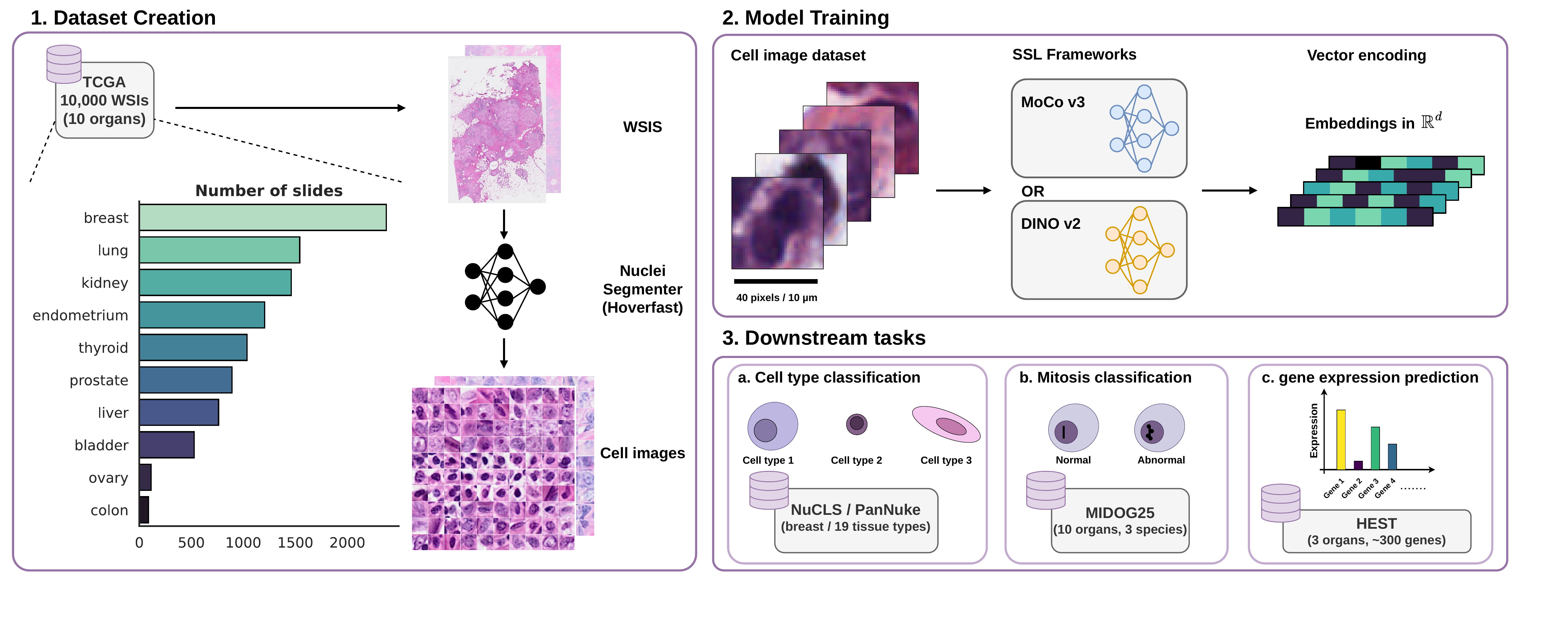}
    \caption{Overview of the LEMON framework.}
    \label{fig:lemon}
\end{figure}

\subsection{A large-Scale Cell Dataset}

To train our models, we generated a large-scale dataset of cell images from \ac{tcga} database \cite{weinstein_cancer_2013}. We processed around 10,000 \acp{wsi} scanned at 40X resolution spanning 10 distinct cancer types. To perform cell segmentation at this scale, we employed \textbf{HoverFast} \cite{liakopoulos_hoverfast_2024}, a highly optimized implementation of the Hover-Net architecture that provides a 20x speed-up, making slide-level segmentation computationally tractable. For each detected nucleus, we extracted a $40 \times 40$ pixel patch centered on the nucleus centroid. After that we selected random nuclei from each \ac{wsi}. This process yielded a pre-training dataset of several hundred million cell instances. The small patch size, a key characteristic of this domain, contrasts with the larger resolutions (e.g., $224 \times 224$) common in natural images, necessitating adaptations in both model architecture and training methodology.

\subsection{\ac{ssl} for cell images}
\label{sec:ssl_cell_images}

We adapted and compared the two leading \ac{ssl} paradigms, contrastive learning (\textbf{MoCov3} \cite{chen_empirical_2021}) and non-contrastive learning (\textbf{DINOv2} \cite{oquab_dinov2_2024}), tailoring them to the specific properties of our cell image data.

\paragraph{Architectural and Framework Adaptations.}
Given the small input resolution of our cell patches, we utilized smaller model backbones than those typically used for \ac{ssl} in natural images. For the contrastive approach with MoCov3, we experimented with \ac{vit} \cite{dosovitskiy_image_2021}, \ac{vit}s refers to the \textit{small} architecture and \ac{vit}b to the \textit{base} architecture. We used a patch size of 8 pixels in our experiments which we indicate with \ac{vit}s8 for a \textit{small} architecture. For the DINOv2 framework, we made two critical modifications to the loss function. First, we removed the local-to-global matching component of the DINO loss, as local crops of an already small $40 \times 40$ patch carry insufficient information to be meaningful. Second, consistent with observations in other pathology applications \cite{zimmermann_virchow2_2024}, we found the effect of the KoLeo regularizer to be excessively \textit{strong}, potentially leading to numerical errors in the loss. We therefore replaced it with a \textbf{more moderate} \ac{kde} based loss, which encourages global feature uniformity without excessively penalizing local neighborhoods in the embedding space:

\begin{equation}
    \mathcal{L}_{\text{KDE}} = \sum_z \log \sum_{z'} k(z, z')
\end{equation}
where $k$ is a kernel function measuring similarity between embeddings.

\paragraph{Domain-Specific Data Augmentations.}
We designed two custom augmentations to address the unique characteristics of cellular images. First, to enforce rotational invariance, a critical prior for cell morphology, which has no canonical orientation, we implemented a full \textbf{random rotation} from $0^\circ$ to $360^\circ$ similar to \cite{alfasly_rotation-agnostic_2024}. To avoid edge artifacts (e.g., black corners), we implemented this by extracting a larger $60 \times 60$ patch, performing the rotation, and then taking a central $40 \times 40$ crop. Second, we evaluated a \textbf{random stain augmentation} \cite{shen_randstainna_2022} based on Macenko's method \cite{macenko_method_2009}. This simulates the variability in \ac{he} staining that arises from different slide preparation protocols, training the model to learn features that are invariant to color and intensity shifts and thus more robust for downstream applications.

\subsection{Benchmarking datasets}

To comprehensively evaluate the quality of the learned cell representations, we benchmarked our models on nine distinct downstream tasks, encompassing six classification tasks and three regression tasks on various types. These benchmarks were selected to test various aspects of the embeddings, including fine-grained cell type discrimination, generalization across different domains, and the ability to predict molecular features from morphology.

\begin{itemize}
    \item \textbf{NuCLS} \cite{amgad_nucls_2022}: A multi-class classification dataset of single-cell images from breast cancer \ac{he} slides. Its key feature is a hierarchical annotation scheme, with labels reviewed and confirmed by pathologists. This structure allows us to evaluate the representative power of our embeddings on both fine-grained (e.g., distinguishing immune cell types) and coarse-grained (e.g., tumor vs. stromal) classification tasks. We refer to them as NuCLS (super) for broad categories, NuCLS (main) for main cell type categories and NuCLS (raw) for fine-grained type and subtype classifications.  It is derived from TCGA WSIs and can be considered an in-domain dataset for LEMON.

    \item \textbf{PanNuke} \cite{gamper2020pannuke}: A large-scale, multi-organ dataset of segmented and labeled nuclei from diverse tissue types. It contains a broad range of nuclear morphologies and cell types across multiple cancer and non-cancer tissues, making it well suited for evaluating generalization across organs and histological contexts. Here, we use PanNuke for single-cell classification, assessing the robustness of learned representations in a heterogeneous, multi-domain setting.  It is primarily an \ac{ood} dataset for LEMON, although some images may originate from TCGA.
    
    \item \textbf{MIDOG25} (track 2, training dataset)\cite{weiss_dataset_2025}: A dataset for binary classification of mitotic figures (atypical vs. normal). It is specifically designed to test model robustness and generalization, as it contains cell images sourced from multiple species (human and canine), diverse organs, and various digital slide scanners. It is an \ac{ood} dataset for LEMON.
    
    \item \textbf{HEST-derived Gene Expression Regression} \cite{jaume_hest-1k_2024}: To assess the capacity of our embeddings to capture morpho-molecular relationships, we curated a novel regression dataset from the HEST database. We created 3 datasets by using slides originating from breast cancer (HEST-breast), lung pulmonary fibrosis (HEST-lung) and bowel cancer (HEST-bowel). It is an \ac{ood} dataset for LEMON.

\end{itemize}

Further details on each dataset and preprocessing are provided in the supplementary materials.

\subsection{Embedding evaluation}

To assess the quality and discriminative power of the features learned by our \ac{ssl} models, we followed the standard \textbf{linear probing} protocol. This evaluation involves training a simple linear model on fixed embeddings obtained from the pretrained model to perform a specific downstream task. The rationale for this approach is that a powerful feature extractor should produce representations that are linearly separable. Consequently, the performance of a simple linear model serves as a direct and reliable proxy for the quality of the learned embeddings. Alternative protocols like full fine-tuning, parameter-efficient fine-tuning or few shot learning also exist, but these methods generally produce similar conclusions about relative model performance \cite{marks_closer_2024}. In our experiments, we trained a \textbf{logistic regression} classifier on the embeddings for the classification tasks (NuCLS, PanNuke and MIDOG25) and a \textbf{ridge regression} model for the gene expression prediction task (HEST). The performance of these linear models is used both to directly compare our approach against competing methods and to motivate key design choices, including methodological decisions and the composition of the pretraining dataset.

Second, we conducted an additional comparative analysis to assess how different representations behave under domain shift. To this end, we simulated staining variations \cite{macenko_method_2009}, and evaluated the changes induced in the latent space across the different SSL strategies. In addition, we measured the impact of this staining-induced domain shift on downstream classification performance on NuCLS and MIDOG25.

Third, we evaluated how well our embeddings capture biological differences by comparing the morphology-driven structure of the latent space with the expression patterns of well-known marker genes.

\section{Results}

\subsection{Experimental setup}

Unless stated otherwise, we used a dataset of 10M cell images from 10 organs. All the models using the MoCo v3 pretraining strategy are trained for 150 epochs with an epoch length of 1M images. The batch size is set to 4096. For the augmentations, we used the best-performing configuration identified through ablation (see \autoref{app-domain_aug}). 

\subsection{Full benchmarking}

To assess performance, we benchmarked our model against strong baselines based on pre-trained representations, organized into four families: (1) transfer learning from ImageNet \cite{russakovsky_imagenet_2015}; (2) \ac{ssl} approaches trained on natural images such as DINOv2 \cite{oquab_dinov2_2024}; (3) histopathology foundation models pre-trained on \ac{wsi} tiles, including UNI \cite{chen_towards_2024}, H-Optimus-0 \cite{saillard_h-optimus-0_2024}, and Virchow2 \cite{zimmermann_virchow2_2024}; and (4) cell-level representations, including Volta \cite{nakhli_volta_2024}, the current state of the art. For transfer learning \cite{zhuang_comprehensive_2020}, we extracted embeddings from each model’s penultimate layer. Because most baselines (except Volta) require 224×224 inputs, we resized nuclei images to the correct size. Results are reported in \autoref{tab:bench}.

Our method decisively outperformed models pre-trained on natural images across all datasets. On MIDOG25, for instance, it achieves a balanced accuracy of $0.746 \pm 0.015$, compared with $0.604 \pm 0.005$ for an ImageNet-pretrained ResNet-50 and $0.654 \pm 0.007$ for the best DINOv2 variant. These gaps underline the limits of natural-image features and the importance of domain-specific pre-training for cellular morphology.

Against contemporary tile-level \acp{fm} (UNI, H-Optimus-0, Virchow2), our approach performs significantly better on MIDOG25 and NuCLS, and remains highly competitive on PanNuke and HEST gene-expression prediction tasks. Gains are particularly pronounced on MIDOG25, which depends on fine-grained cellular cues that tile-level \ac{fm}s tend to overlook. In order to check that this result is not entirely due to domain shift introduced by feeding nucleus images to tile-level \ac{fm}, we further investigated alternative strategies to extract cell-level encodings from tile-level \ac{fm} applied to larger fields of view matching the \ac{fm} training distribution (see below).

Compared to Volta, the only other cell-level baseline, our model held a clear advantage across all tasks. It surpassed Volta on MIDOG25, NuCLS and PanNuke classification tasks and delivered stronger results on HEST gene-expression prediction, indicating more robust and generalizable cellular representations for downstream computational pathology.

Additional evaluation metrics for MIDOG25, NuCLS and PanNuke are provided in the supplementary materials.

\begin{table}[!hb]
\caption{Performance of pretraining strategy on downstream tasks. We report balanced accuracy for MIDOG25 (majority), NuCLS (super) and PanNuke (classification) and \ac{pcc} for HEST (breast, lung, bowel). Values are mean (s.d.) over folds, best values are highlighted in bold.}
\label{tab:bench}
\centering
\resizebox{\textwidth}{!}{%
\begin{tabular}{lcccccc}
\toprule
 & MIDOG25 & NuCLS & PanNuke & HEST-Bowel & HEST-Breast & HEST-Lung \\
 & bal acc & bal acc & bal acc & pcc & pcc & pcc \\
Embeddings &  &  &  &  &  &  \\
\midrule

\multicolumn{7}{@{}l}{\textit{Natural images TL}}\\
Imagenet (ResNet-18) & 0.577 (0.008) & 0.693 (0.005) & 0.56 (0.004) & 0.114 (0.023) & 0.113 (0.023) & 0.083 (0.021) \\
Imagenet (ResNet-50) & 0.603 (0.012) & 0.711 (0.005) & 0.587 (0.005) & 0.12 (0.035) & 0.124 (0.023) & 0.091 (0.013) \\
\midrule

\multicolumn{7}{@{}l}{\textit{Natural images FM}}\\
DINOv2 (ViT-s/14) & 0.643 (0.004) & 0.732 (0.003) & 0.635 (0.009) & 0.146 (0.016) & 0.157 (0.029) & 0.116 (0.028) \\
DINOv2 (ViT-b/14) & 0.653 (0.013) & 0.732 (0.003) & 0.627 (0.01) & 0.145 (0.022) & 0.156 (0.026) & 0.118 (0.028) \\
DINOv2 (ViT-l/14) & 0.647 (0.01) & 0.721 (0.004) & 0.619 (0.008) & 0.142 (0.016) & 0.153 (0.026) & 0.117 (0.029) \\
DINOv2 (ViT-g/14) & 0.654 (0.01) & 0.733 (0.008) & 0.631 (0.007) & 0.153 (0.021) & 0.164 (0.025) & 0.118 (0.029) \\
\midrule

\multicolumn{7}{@{}l}{\textit{Tile level FM}}\\

UNI (ViT-l/16) & 0.623 (0.008) & 0.733 (0.007) & 0.643 (0.008) & 0.15 (0.035) & 0.151 (0.029) & 0.118 (0.029) \\
Virchow2 (ViT-h/14) & 0.618 (0.016) & 0.753 (0.008) & 0.645 (0.009) & 0.148 (0.034) & 0.154 (0.029) & 0.11 (0.029) \\
H-Optimus-0 (ViT-g/14) & 0.631 (0.012) & 0.752 (0.006) & \textbf{0.648} (0.007) & 0.159 (0.029) & \textbf{0.157} (0.029) & 0.119 (0.029) \\
\midrule

\multicolumn{7}{@{}l}{\textit{Cell level FM}}\\
Volta (ResNet-18) & 0.726 (0.01) & 0.736 (0.003) & 0.571 (0.007) & 0.148 (0.025) & 0.115 (0.028) & 0.118 (0.025) \\
LEMON-MoCov3 (ViT-s/8) & \textbf{0.741} (0.02) & \textbf{0.779} (0.005) & 0.641 (0.008) & \textbf{0.16} (0.027) & 0.156 (0.021) & \textbf{0.128} (0.028) \\
\bottomrule
\end{tabular}
}
\end{table}

\subsection{Comparison of cell embedding strategies from tile-level \ac{fm}s}

Single cells correspond to regions of roughly $40\times40$ pixels. To use tile-level \acp{fm} trained on $224\times224$ (tiles) for single-cell representation, the standard approach is to resize nuclei images to $224\times224$, which may deviate substantially from the original training distribution. Here, we investigate the direct use of \ac{fm} token encodings as an alternative. 

For this, we focus on MIDOG25, as for this task, the cell labels are independent from spatial context. We extracted $128\times128$ crops at $40\times$ magnification centered on each nucleus and resized them to $224\times224$ to match the expected input size. Because we did not have access to the full slides, using $224\times224$ crops directly would have led to many exclusions due to border effects. With this procedure, we retained $9{,}115$ nuclei out of the original $11{,}939$. The label distribution is provided in the supplementary material.

To obtain embeddings from tile-level \ac{fm}, we investigated two approaches to derive the embeddings used in the downstream task. In the \textbf{class token} approach, we used the class token embedding as the representation. In the \textbf{center tokens} approach, we used the mean embedding of the $4\times 4$ centered patch tokens of the full image. With our setting, these patches correspond to roughly $8~\mu m\times 8~\mu m$ in the center of the images, which should contain all the relevant information for the mitosis classification task.

We report the results in \autoref{tab:midog25_spatial_bench}. Across all tile-level \acp{fm}, patch tokens consistently outperform the class token approach. Our model surpasses all competing methods, further confirming the relevance of our cell-level \ac{fm} for cell-level tasks.

\begin{table}[!hb]
\caption{Evaluation of representations from larger images ($128\times128$ pixel crop size) with patch level foundation models on the MIDOG25 dataset}
\label{tab:midog25_spatial_bench}
\centering
\begin{tabular}{lll}
\toprule
Target & \multicolumn{2}{c}{MIDOG25 (majority)} \\
Metric & bal acc & f1 (weighted) \\
Embeddings &  &  \\
\midrule
\textbf{UNI-class (ViT-l/16)} & 0.539 (0.007) & 0.834 (0.004) \\
\textbf{UNI-center (ViT-l/16)} & 0.645 (0.019) & 0.876 (0.007) \\
\midrule
\textbf{Virchow2-class (ViT-h/14)} & 0.538 (0.005) & 0.833 (0.002) \\
\textbf{Virchow2-center (ViT-h/14)} & 0.673 (0.017) & 0.886 (0.006) \\
\midrule
\textbf{H-Optimus-0-class (ViT-g/14)} & 0.524 (0.009) & 0.826 (0.005) \\
\textbf{H-Optimus-0-center (ViT-g/14)} & 0.644 (0.011) & 0.875 (0.004) \\
\midrule
\textbf{LEMON-MoCov3 (ViT-s/8)} & 0.727 (0.02) & 0.905 (0.007) \\
\bottomrule
\end{tabular}

\end{table}

\subsection{\ac{flops} comparison}

Computational efficiency is crucial in the single-cell setting, where a single whole-slide image may contain millions of nuclei, making large-scale embedding extraction prohibitively expensive with heavyweight architectures. Tile-level foundation models typically rely on very large backbones (e.g., ViT-H/14, ViT-g/14) with high computational footprints, whereas LEMON leverages a lightweight ViT-S/8 backbone, operating with orders of magnitude fewer \ac{flops}.

\begin{figure}
    \centering
    \includegraphics[width=\linewidth]{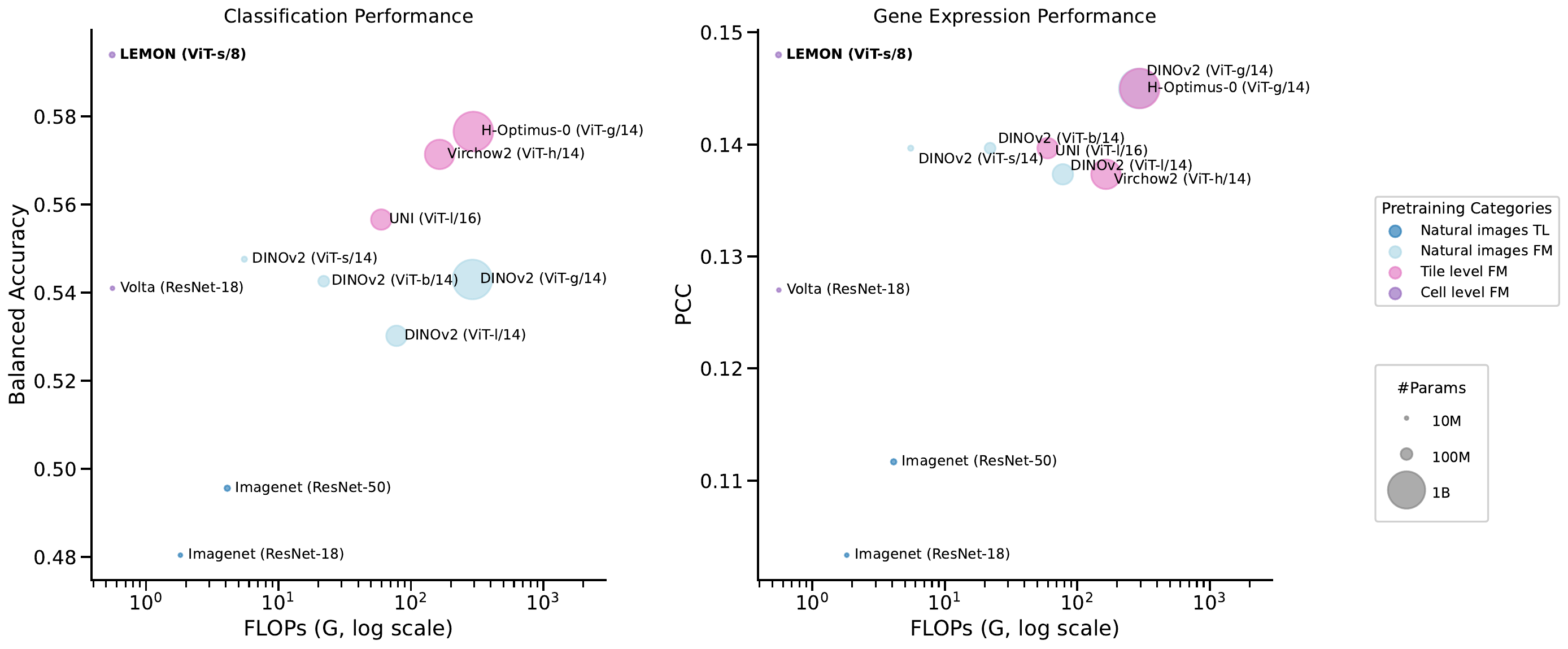}
    \caption{\textbf{Performance of pretraining strategies versus \ac{flops}.} Left: mean balanced accuracy across NuCLS (super, main, raw), MIDOG25 and PanNuke for classification. Right: mean \ac{pcc} across breast, lung, and bowel datasets for regression. Marker size indicates model parameter count; color indicates pretraining category.}
    \label{fig:flop}
\end{figure}

To evaluate the trade-off between performance and computational cost, we compared different pretraining strategies by plotting downstream performance against the number of \ac{flops} required during inference (\autoref{fig:flop}).
Our cell-level models achieve state-of-the-art performance while striking an optimal balance between representational power and efficiency.

Histology \ac{fm}s outperform natural image pretraining at comparable \ac{flops}. 
Interestingly, for classification tasks, histopathology tile-level models generally exhibit improved performance as model size and \ac{flops} increase. In contrast, LEMON breaks this trend, achieving the highest overall accuracy (NuCLS super: 0.779; MIDOG25: 0.741) with only 0.5G \ac{flops}, compared to 295.9G \ac{flops} for H-Optimus-0, the second-best performing model. For gene expression prediction, a more challenging task, the overall trends remain consistent.

\subsection{Choice of \ac{ssl} framework}

We compared two state-of-the-art \ac{ssl} methods, MoCov3 \cite{chen_empirical_2021} and DINOv2 \cite{oquab_dinov2_2024}, as pretraining strategies for cell patches, given their strong performance in tile-level pathology foundation models.
While DINOv2 is among the most powerful SSL frameworks, we observed limitations when applied to small 40×40 images. To address these, we introduced - as detailed in section \ref{sec:ssl_cell_images} - two modifications to the loss function: (i) removing the local-to-global matching term, which is ill-suited for such small patches, and (ii) replacing the strong KoLeo regularizer with a milder \ac{kde}-based loss. These modifications improved performance across all classification benchmark datasets (see \autoref{app-compssl}). 

Despite these adjustments, MoCov3 consistently outperformed DINOv2, suggesting that contrastive \ac{ssl} is more robust in low-resolution pathology scenarios. We hypothesize that DINOv2’s strength largely stems from its challenging local–global objective, which enforces structural understanding but becomes ill-posed for very small patches, thereby limiting its effectiveness and favoring the global contrastive objective of MoCov3.
Consequently, we focused the remainder of the study on MoCov3, and LEMON refers to LEMON-MoCov3.

\subsection{Dataset Composition}

\begin{figure}
    \centering
    \includegraphics[width=\linewidth]{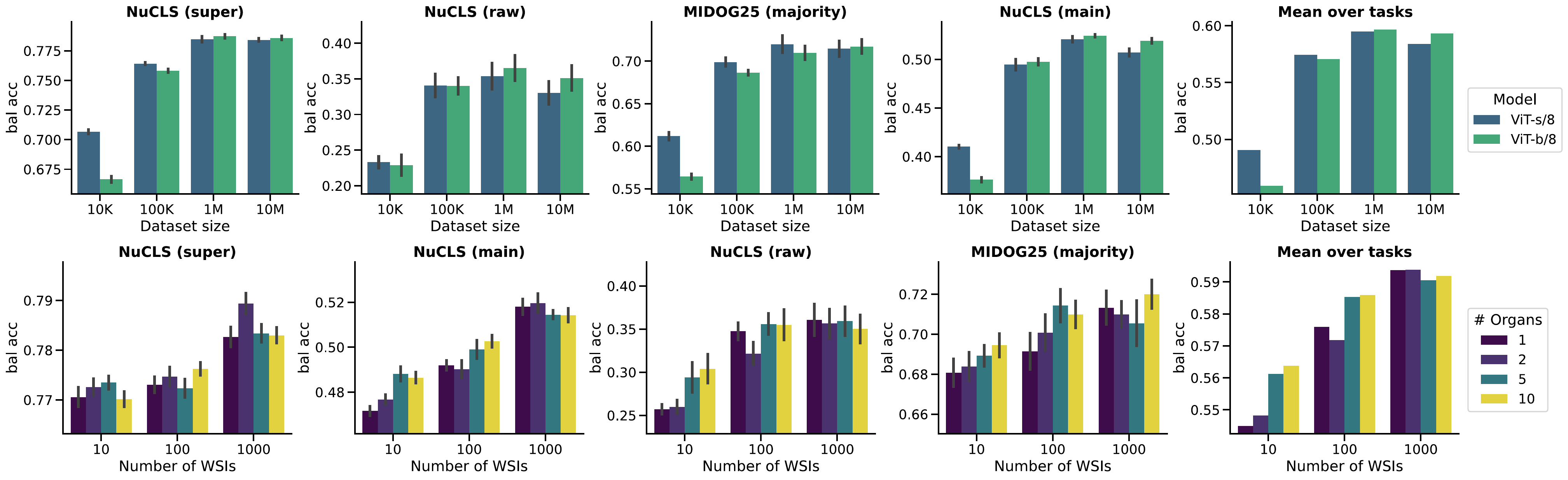}
    \caption{\textbf{Performance of models vs. training data composition.} Mean balanced accuracy (error bars represents standard errors) across MIDOG25 and the three NuCLS classification tasks. \textit{Top:} Performance by training dataset size; models improve with larger datasets and level off near 1M images. \textit{Bottom:} With the total number of images held constant, models perform better as diversity increases (more organs of origin and more slides).}
    \label{fig:data}
\end{figure}

We first examined how dataset scale affects benchmark classification. As shown in the top row of \autoref{fig:data}, balanced accuracy increases consistently as the dataset grows, before saturating around 1M images for both \ac{vit}-S/8 and \ac{vit}-B/8. We kept \ac{vit}-S/8 as increasing the model sizes did not lead to significantly better results with our training scheme.

We then isolated the role of data diversity, motivated by evidence that training set diversity can strongly influence downstream performance \cite{oquab_dinov2_2024, vo_automatic_2024}. Fixing the dataset size to 1M images, we varied the number of \acp{wsi} and the number of source organs from which cell images were extracted. To select representative slides, we used a K-medoids strategy based on TITAN-derived embeddings \cite{ding_multimodal_2025}. As shown in the bottom row of \autoref{fig:data}, gains are driven primarily by increasing the number of slides, while performance remained relatively stable with respect to the number of organs, provided that the number of slides is sufficiently high. This suggests the model does not require organ-specific pretraining for effective application across organs. 

\subsection{Robustness to Staining Variation}

Histopathology images exhibit substantial staining variability across acquisition centers, inducing embedding shifts unrelated to biological content. Learning stain-invariant representations is therefore essential and strongly depends on training-time color augmentations. To assess staining robustness, we applied Macenko-based stain normalization using six reference images from TCGA breast samples spanning six distinct centers (see \autoref{app-stainingrobustness}).

We encoded MIDOG25 under each reference stain and quantified embedding shifts using root mean squared error (RMSE), cosine similarity, and k-nearest neighbor overlap (k=100), averaged across stains. Metrics were computed in PCA space after L2 normalization to ensure a fair comparison across models. Results are reported in \autoref{tab:macenko}. LEMON shows the highest stability across all metrics. Volta exhibits intermediate robustness, while H-Optimus-0 is the most sensitive to staining variation, demonstrating larger embedding shifts than the cell-level encoders.

Importantly, this robustness does not compromise discriminative power: under all six staining conditions on MIDOG25 and NuCLS, LEMON still consistently outperforms both Volta and H-Optimus-0 in terms of balanced accuracy (see \autoref{tab:macenko} and supplementary materials).

\begin{table}[!hb]
\caption{Robustness to Macenko stain normalization across 6 TCGA reference images. Embedding robustness metrics measure the shift in embedding space between original and stain-normalized cells. Biological tasks report balanced accuracy on stain-normalized embeddings. Best in bold.}
\label{tab:macenko}
\centering
\begin{tabular}{lccc}
\toprule
  & LEMON-MoCov3 & Volta & H-Optimus-0 \\
  & (ViT-s/8) & (ResNet-18) & (ViT-g/14) \\
\midrule
\multicolumn{4}{@{}l}{\textit{Embedding Robustness}} \\
RMSE (↓) & \textbf{0.777} (0.150) & 0.879 (0.266) & 1.311 (0.076) \\
Cosine Similariy (↑) & \textbf{0.687} (0.116) & 0.579 (0.246) & 0.138 (0.099) \\
Neightborhood Overlap (↑) & \textbf{0.398} (0.067) & 0.297 (0.059) & 0.154 (0.014) \\
\midrule
\multicolumn{4}{@{}l}{\textit{Biological tasks}} \\
MIDOG - bal acc (↑) & \textbf{0.734} (0.011) & 0.699 (0.015) & 0.618 (0.011) \\
NuCLS - bal acc (↑) & \textbf{0.755} (0.006) & 0.718 (0.004) & 0.736 (0.004) \\
\bottomrule
\end{tabular}
\end{table}

\subsection{Latent space organization reflects biological identity}

\begin{figure}
    \centering
    \includegraphics[width=\textwidth]{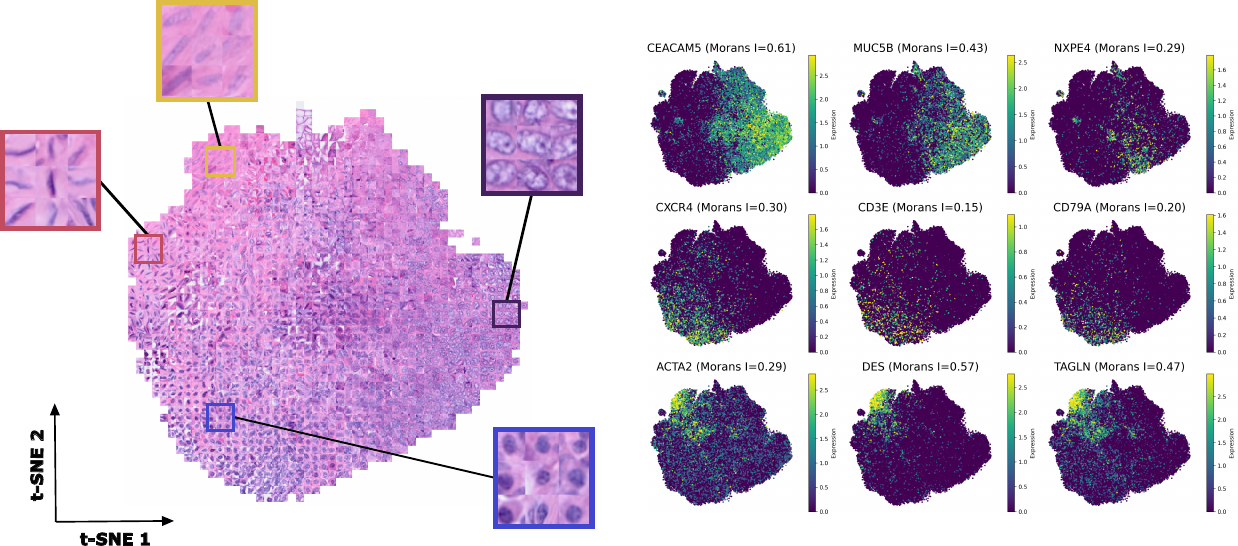}
    \caption{Alignment between morphological embeddings and marker-gene expression in a bowel Xenium tissue section (HEST id: TENX147). Left: t-SNE projection of LEMON morphological embeddings. Right: normalized expression maps for marker genes, showing their spatial distribution within the morphology-derived manifold.}
    \label{fig:tsne-xenium}
\end{figure}

We examined whether the learned cell embeddings capture biological structure by projecting them into two dimensions using t-SNE (\autoref{fig:tsne-xenium}). The latent space exhibited a coherent and continuous organization: epithelial cells were predominantly located on the right, while the left side displayed structured variation, with elongated cells in the upper region and smaller round-nuclei cells in the lower region.

To assess biological relevance, we compared the organization of the embedding space with the expression of marker genes associated with major cell types (epithelial, stromal, and immune) in a Xenium tissue section. Although the model was not trained on gene expression data, marker expression concentrated in distinct regions of the embedding space (\autoref{fig:tsne-xenium}). We quantified this correspondence using Moran’s I computed on the k-nearest-neighbour graph in embedding space, which measures autocorrelation of expression values among neighboring cells. All markers showed positive autocorrelation, and genes associated with the same cell types exhibited similar spatial patterns, indicating that transcriptionally related cells occupy nearby regions in the morphology-derived manifold.

This analysis can be extend to any whole-slide images. On a TCGA sample, independent pathologist annotations confirmed the coherence of the latent space and its alignment with cell types (see \autoref{app-representationanlysis}).

\section{Conclusion}
 
In this paper, we introduce LEMON, a self-supervised model for nuclei image representation learning. We build a large-scale dataset of nucleus crops and show that our approach achieves state-of-the-art performance across five downstream tasks. Notably, we demonstrate that self-supervised learning can be effectively applied to very small images (40×40) with minor adaptations to existing frameworks. By carefully tuning training schemes and augmentations, LEMON produces stable and meaningful embeddings that capture nuclear morphology and chromatin organization, reducing the need for manual labels and enabling new explorations of the morphological landscape of cell images. Furthermore, we show that our representations are robust to staining-induced domain shifts and truly capture biological signals, as they correlate with the expression of well-known marker genes.
Finally, we make the dataset of all nuclei detection, the pretext task targets and our foundation model publicly available to the scientific community. Beyond \ac{cp}, these results suggest that \ac{ssl} could be extended to other domains where only small image patches are available.

In future work, an interesting avenue is to compare representations learned at different physical scales (nucleus, tissue and slide level views) to optimally integrate multi-scale information. Finally, our representations provide an excellent starting point for in-depth spatial modeling and tissue architecture analysis. We deliberately chose not to incorporate the spatial distribution of cells into this foundation model, so that future work on spatial analyses can remain disentangled from single-cell morphological representations, which we consider an important asset for future developments.

\section*{Acknowledgements}

This work was funded by the French government under the management of Agence Nationale de la Recherche as part of the “Investissements d’avenir” program, reference ANR-19-P3IA-0001 (PRAIRIE 3IA Institute) and ANR-23-IACL-0008 (PRAIRIE-PSAI). Furthermore, this work was supported by a government grant managed by the Agence Nationale de la Recherche under the France 2030 program, with the reference numbers ANR-24-EXCI-0001, ANR-24-EXCI-0002, ANR-24-EXCI-0003, ANR-24-EXCI-0004, ANR-24-EXCI-0005. This work was granted access to the HPC resources of IDRIS under the allocation 2025-AD011016560 made by GENCI.

\bibliography{references}
\bibliographystyle{unsrtnat}
\clearpage

\appendix

\setcounter{figure}{0}
\setcounter{table}{0}
\renewcommand{\thefigure}{S\arabic{figure}}
\renewcommand{\thetable}{S\arabic{table}}

\section{Benchmarking datasets}\label{app-data}

\paragraph{NuCLS dataset:} a dataset of images from slides originating from breast cancer of annotated cell type images, composed of three classification tasks with increased cell type resolution \textit{super}, \textit{main}, \textit{raw}. The label distribution is shown in \autoref{tab:nucls}.

\begin{table}[!hb]
  \caption{NuCLS dataset with different annotation levels}
  \label{tab:nucls}
  \centering
  \begin{subtable}[t]{0.32\textwidth}
    \centering
    \caption{NuCLS (super)}
    \label{tab:nucls-super}
    \begingroup
      \setlength{\tabcolsep}{3pt}
      \renewcommand{\arraystretch}{0.95}%
      \begin{adjustbox}{width=\linewidth,keepaspectratio}
        \begin{tabular}{ll}
Cell type       & Number of images \\ \hline
tumor\_any      & 16052            \\
sTIL            & 13642            \\
nonTIL\_stromal & 7613            
\end{tabular}

      \end{adjustbox}
    \endgroup
  \end{subtable}\hfill
  \begin{subtable}[t]{0.32\textwidth}
    \centering
    \caption{NuCLS (main)}
    \label{tab:nucls-main}
    \begingroup
      \setlength{\tabcolsep}{3pt}
      \renewcommand{\arraystretch}{0.95}
      \begin{adjustbox}{width=\linewidth,keepaspectratio}
        \begin{tabular}{ll}
Cell type            & Number of images \\ \hline
tumor\_nonMitotic    & 15874            \\
lymphocyte           & 9617             \\
nonTILnonMQ\_stromal & 6634             \\
plasma\_cell         & 4025             \\
macrophage           & 979              \\
tumor\_mitotic       & 178             
\end{tabular}

      \end{adjustbox}
    \endgroup
  \end{subtable}\hfill
  \begin{subtable}[t]{0.32\textwidth}
    \centering
    \caption{NuCLS (raw)}
    \label{tab:nucls-raw}
    \begingroup
      \setlength{\tabcolsep}{3pt}
      \renewcommand{\arraystretch}{0.95}
      \begin{adjustbox}{width=\linewidth,keepaspectratio}
        \begin{tabular}{ll}
Cell type             & Number of images \\ \hline
tumor                 & 15874            \\
lymphocyte            & 9617             \\
fibroblast            & 6258             \\
plasma\_cell          & 4025             \\
macrophage            & 979              \\
ductal\_epithelium    & 399              \\
vascular\_endothelium & 376              \\
apoptotic\_body       & 293              \\
mitotic\_figure       & 178              \\
neutrophil            & 35               \\
myoepithelium         & 33               \\
eosinophil            & 2               
\end{tabular}

      \end{adjustbox}
    \endgroup
  \end{subtable}
\end{table}

\paragraph{MIDOG25 dataset:} a dataset of images from slides originating from multiple organs and species with mytotic figure annotations. We did not use any of the metadata to create our splits except for labels to generate stratified splits. The label distribution is shown in \autoref{tab:midog}. 

\begin{table}[!hb]
\caption{MIDOG25 dataset.}
\label{tab:midog}
\centering
\begin{tabular}{ll}
Class           & Number of images \\ \hline
NMF             & 10191            \\
AMF             & 1748            \\
\end{tabular}
     
\end{table}

For the comparison of LEMON to tile-level \ac{fm} on larger images (matching tile tile-level input dimension), we extracted crops of size $128\times 128$ pixels at 40X magnification, centered on each nucleus, which were then resized to $224\times 224$ to match the expected image size.  We selected a crop size of  $128\times 128$ pixels to avoid border effects, as the MIDOG25 dataset did not consist of full WSI. This size therefore allowed us to maximize the number of usable nuclei while remaining compatible with the image distribution used to train the tile-level \ac{fm} models.
With this setting, we retained $9115$ nuclei out of the original $11939$. The label distribution can be seen in \autoref{tab:midog_spatial}.

\begin{table}[!hb]
\caption{MIDOG25 dataset with $128\times128$ crop images}
\label{tab:midog_spatial}
\centering
\begin{tabular}{ll}
Class           & Number of images \\ \hline
NMF             & 7955            \\
AMF             & 1160            \\
\end{tabular}
\end{table}

\paragraph{PanNuke dataset:} a dataset of images from slides originating from multiple organs and species with broad cell type annotations. We extracted cell images for cells whose associated images of size 60 by 60 fall within the full crops with the annotations to make sure that all the part of the cell is in the images. It resulted in the following dataset. The dataset was originally published in \cite{gamper2020pannuke}, we downloaded the data from \url{https://huggingface.co/datasets/RationAI/PanNuke}. The classes are 0: Neoplastic, 1: Inflammatory, 2: Connective, 3: Dead and 4: Epithelial.

\begin{table}[!hb]
\caption{PanNuke dataset.}
\label{tab:pannuke}
\centering
\begin{tabular}{ll}
Class & Number of images \\ \hline
0     & 39222            \\
1     & 26108            \\
2     & 17258            \\
3     & 13457            \\
4     & 1582            
\end{tabular}

\end{table}

\paragraph{HEST dataset:}
This dataset is built from Xenium slides, a technology for \ac{ist} that measures up to hundreds of distinct \ac{mrna} species while preserving their spatial localisation. After cell segmentation, we construct cell–count matrices in which each row corresponds to a cell, each column to a gene, and each entry to the number of transcripts detected in that cell. We use the segmentation provided by the Xenium software. The platform also provides a co-registered \ac{he} slide aligned with the cell–count matrix, enabling extraction of cell images that are spatially aligned with the transcriptomic measurements.

The prediction task is to identify the top 50 highly expressed genes, defined as follows. For every slide, we compute the mean expression per gene, rank genes within the slide, combine the ranks across slides, and retain the 50 highest-ranked genes. To standardize sample size, we subsample 10\,000 cells per slide uniformly at random. We report performances with a leave-one-out slide evaluation strategy. 

We accessed the Xenium slides through the HEST database \cite{jaume_hest-1k_2024}. The slide IDs used for each subset are:

\begin{itemize}
    \item \textbf{HEST-bowel}: TENX149, TENX148, TENX147, TENX139, TENX111
    \item \textbf{HEST-breast}: NCBI783, NCBI785, TENX94, TENX95, TENX98
    \item \textbf{HEST-lung}: NCBI856, NCBI857, NCBI858, NCBI859, NCBI860, NCBI861, NCBI864, NCBI865, NCBI866, NCBI867, NCBI870, NCBI873, NCBI875, NCBI876, NCBI879, NCBI880, NCBI881, NCBI882, NCBI883, NCBI884
\end{itemize}

We note that LEMON was pretrained on whole-slide images originating from TCGA. Consequently, NuCLS, which is derived from TCGA WSIs, can be considered an in-domain dataset for LEMON. PanNuke aggregates data from multiple sources, including TCGA. Therefore, it can be seen primarily as an \ac{ood} dataset for LEMON, although a subset of the data may be in-domain. Conversely, MIDOG25, and HEST do not overlap with LEMON’s pretraining sources and can therefore be seen as \ac{ood}.

\section{Comparison of SSL Methods}\label{app-compssl}

We evaluated two state-of-the-art \ac{ssl} methods, MoCo v3 and DINOv2, on small cell histopathology patches (40×40). As discussed in section 2.2, we tested several variants of DINOv2, including the original baseline and modified versions where (i) the local-to-global matching term was removed, and (ii) the KoLeo regularizer was replaced with a KDE-based loss.

The KDE loss employs a von Mises–Fisher kernel:
\begin{equation}
    k_{vMF}(x,y)=exp(\kappa x^T y)
\end{equation}
where $\kappa$ controls the concentration of the kernel around the mean direction. In our experiments, we set $\kappa$ to 5 and KDE loss weight to 0.05. This encourages global feature uniformity while avoiding excessive penalties on local neighborhoods.

\autoref{tab:dinov2} summarizes the performance of MoCo v3 and DINOv2 variants across the five benchmark datasets. MoCov3 consistently achieves the best results, outperforming all DINOv2 variants on NuCLS (0.779 vs 0.767), MIDOG (0.746 vs 0.670–0.678), and HEST Bowel (0.160 vs 0.117–0.142), which demonstrates its robustness for small 40×40 patches. The ablations on DINOv2 indicate incremental improvements: removing the local-to-global matching term slightly improves  performances on most dataset (e.g,: Nucls: 0.703 vs 0.681, HEST-Breast 0.129 vs 0.119). Replacing the KoLeo regularizer with the KDE-based loss yields the best DINOv2 variant, improving NuCLS (0.767), HEST Breast (0.143),  HEST Lung (0.147) and HEST Bowel (0.142) relative to the baseline.

\begin{table}[!hb]
\caption{Effect of architectural and framework adaptations on downstream tasks. We report balanced accuracy for NuCLS (super) and MIDOG25 (majority) and PCC for HEST datasets. Values are mean (s.d.), best values are highlighted in bold.}
\label{tab:dinov2}
\centering
\resizebox{\textwidth}{!}{%
\begin{tabular}{lccccc}
\toprule
 & MIDOG25 & NuCLS & HEST-Bowel & HEST-Breast & HEST-Lung \\
 & bal acc & bal acc & pcc & pcc & pcc \\
Model &  &  &  &  &  \\
\midrule

DINOv2-Baseline (vits8) 
& 0.678 (0.011) 
& 0.681 (0.002) 
& 0.117 (0.014) 
& 0.119 (0.023) 
& 0.106 (0.015) \\

DINOv2-No local-to-global (vits8) 
& 0.673 (0.012) 
& 0.703 (0.003) 
& 0.131 (0.016) 
& 0.129 (0.021) 
& 0.110 (0.018) \\

DINOv2-No local-to-global + KDE (vits8) 
& 0.67 (0.011) 
& 0.767 (0.007) 
& 0.142 (0.024) 
& 0.143 (0.024) 
& \textbf{0.147} (0.02) \\

MoCov3 (vits8) 
& \textbf{0.741} (0.02) 
& \textbf{0.779} (0.005) 
& \textbf{0.16} (0.027) 
& \textbf{0.156} (0.021) 
& 0.128 (0.028) \\

\bottomrule
\end{tabular}
}
\end{table}

\section{Augmentation strategies}\label{app-augmentation}

\begin{figure}
    \centering
    \includegraphics[width=\textwidth]{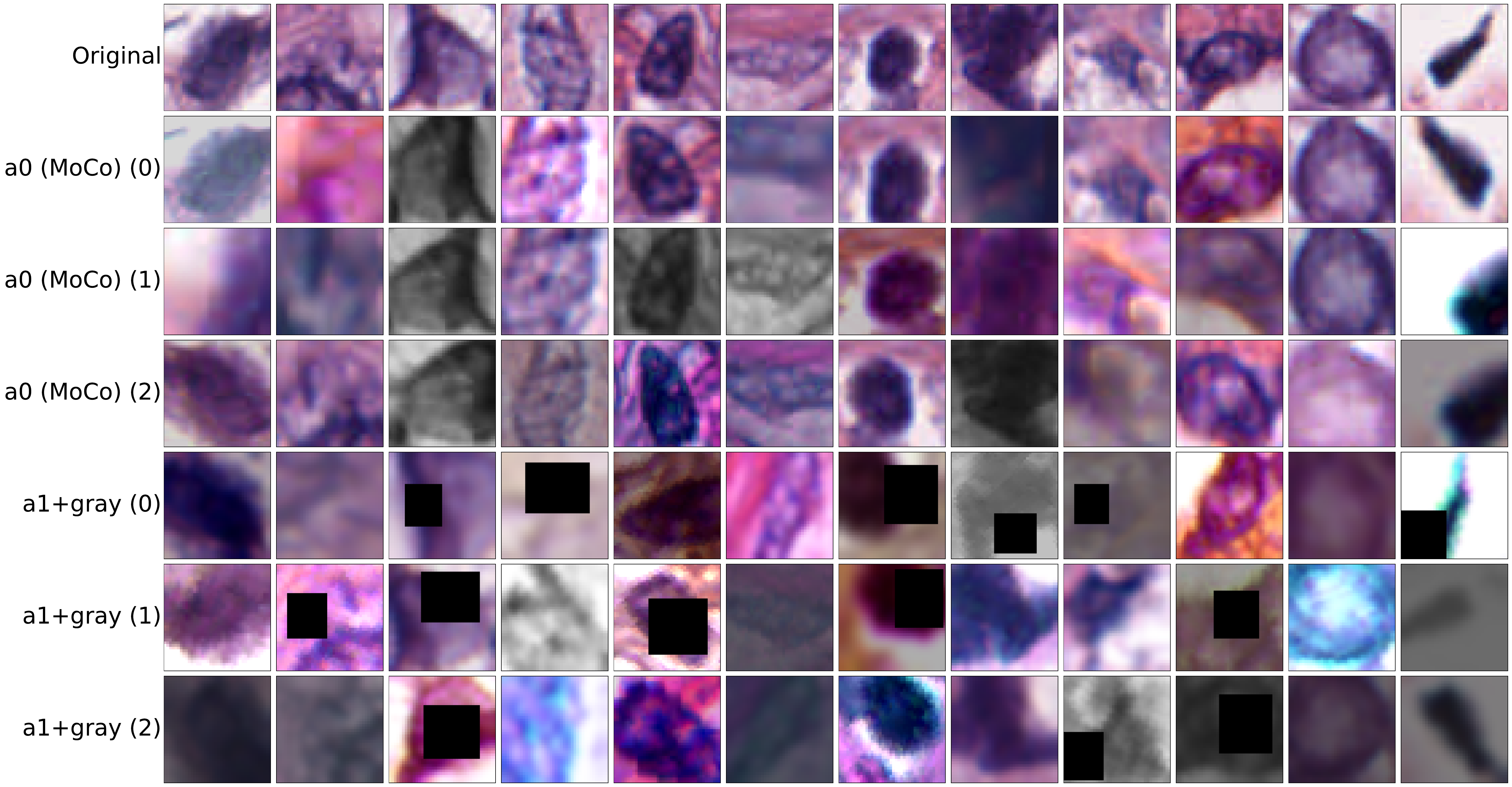}
    \caption{Examples of nuclei images with different augmentation strategies with three random augmentations from either the MoCo v3 augmentation or ours}
    \label{supfig:aug}
\end{figure}

We here detail the augmentation strategies that we used to train our models. 
For MoCov3, we used the augmentations that we found in the pytorch implementation \url{https://github.com/facebookresearch/moco-v3}. We  used torchvision \cite{maintainers_torchvision_2016} implementations.
For \texttt{a0 (MoCo)}:
\begin{itemize}
    \item \texttt{RandomResizedCrop} with scale=(0.08, 1.0)
    \item \texttt{RandomHorizontalFlip}
    \item \texttt{ColorJitter} with parameters = (0.4, 0.4, 0.2, 0.1) and p=0.8
    \item \texttt{RandomGrayscale} with p=0.2
    \item \texttt{GaussianBlur} with sigma=(0.1, 2.0)
\end{itemize}

\section{Domain-specific augmentations}\label{app-domain_aug}

Augmentation policy strongly affects \ac{ssl} performance \cite{morningstar_augmentations_2024}. We revised the pipeline with stronger color jitter, random patch masking, and a customized rotation protocol without padding (\texttt{a1}), which improved both classification (NuCLS, MIDOG25) and gene-expression prediction (HEST) with respect to the standard augmentation scheme (\texttt{a0 (MoCo)}).

Imposing robustness to color variations is important in digital pathology, as staining protocols and scanners may vary between hostpitals. Adding grayscale augmentations (\texttt{a1+gray}) \cite{kang_benchmarking_2023} yielded the best MIDOG25 score and competitive NuCLS results, though HEST effects were mixed. Surprisingly, RandstainNA color perturbations \cite{shen_randstainna_2022} mimicking realistic color variations helped HEST when using a single color template (\texttt{a1+gmm1}) but reduced classification accuracy, and the multi-template variant (\texttt{a1+gmm10}) underperformed overall.

Balancing these trade-offs, we adopted \texttt{a1+gray} as our default, as it combines the strongest classification gains with good HEST performance. The results are summarized in \autoref{tab:aug}. 

\begin{table}[!hb]
\caption{Effect of augmentation policy on downstream tasks. We report balanced accuracy for NuCLS (supervised) and MIDOG25 (majority), and \ac{pcc} for HEST (breast, lung, bowel). Values are mean (s.d.) over folds, best values are highlighted in bold.}
\label{tab:aug}
\centering
\begin{tabular}{lccccc}
\toprule
 & MIDOG25 & NuCLS & HEST-Bowel & HEST-Breast & HEST-Lung \\
 & bal acc & bal acc & pcc & pcc & pcc \\
Augmentations &  &  &  &  &  \\
\midrule
a0 (MoCo) & 0.682 (0.013) & 0.771 (0.003) & 0.154 (0.03) & 0.148 (0.023) & 0.116 (0.027) \\
a1 & 0.715 (0.021) & \textbf{0.784} (0.003) & 0.161 (0.028) & 0.155 (0.025) & \textbf{0.133} (0.032) \\
a1+gray & \textbf{0.746} (0.015) & 0.779 (0.005) & 0.16 (0.027) & 0.156 (0.021) & 0.128 (0.028) \\
a1+gmm1 & 0.736 (0.018) & 0.768 (0.005) & \textbf{0.164} (0.019) & \textbf{0.162} (0.021) & 0.126 (0.029) \\
a1+gmm10 & 0.725 (0.014) & 0.765 (0.003) & 0.160 (0.027) & 0.156 (0.021) & 0.128 (0.028) \\

\bottomrule
\end{tabular}
\end{table}

For \texttt{a1}:
\begin{itemize}
    \item \texttt{RotationCrop} with degree=360 (our implementation which prevent black pixel padding) 
    \item \texttt{RandomResizedCrop} with scale=(0.32, 1.0)
    \item \texttt{RandomHorizontalFlip}
    \item \texttt{ColorJitter} with parameters = (0.6, 0.7, 0.5, 0.2) and p=0.8
    \item \texttt{RandomGrayscale} with p=0.2
    \item \texttt{RandomErasing} with scale=(0.1, 0.3), ratio=(0.8, 1.2) and p=0.3
    \item \texttt{GaussianBlur} with sigma=(0.1, 2.0)
\end{itemize}

For \texttt{a1+gray}, we added \texttt{RandomGrayscale} with p=0.2 to the \texttt{a1} augmentations. 
For stain-specific augmentations, we first computed stain statistics on the training dataset with either 1 or 10 mixtures and then sample a new staining to augment the images. \texttt{a1-gmm1} and \texttt{a1-gmm10} correspond to \texttt{a1} with these additional augmenations.

\section{Additional Robustness to Staining Variability}\label{app-stainingrobustness}

\begin{figure}
    \centering
    \includegraphics[width=\textwidth]{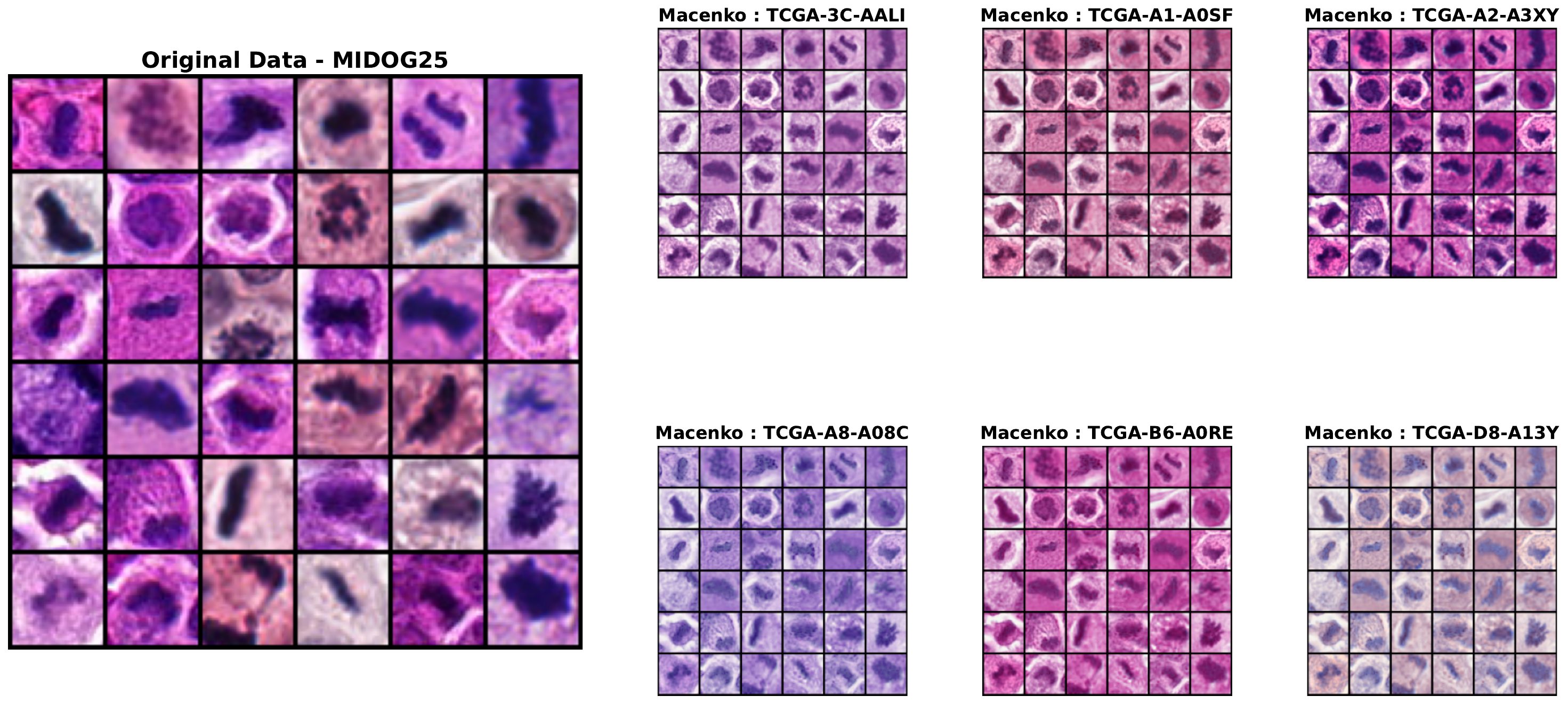}
    \caption{Visualization of Macenko stain augmentation. The first panel shows a subset of original MIDOG25 cell images, while the remaining panels display the same images after Macenko augmentation using six different reference stains.}
    \label{supfig:aug_macenko}
\end{figure}

To provide additional insight into the robustness of learned representations under staining variability, we visualize representative examples of Macenko-based stain normalization applied using six TCGA reference images to MIDOG25 dataset (\autoref{supfig:aug_macenko}).

\begin{figure}
    \centering
    \includegraphics[width=\textwidth]{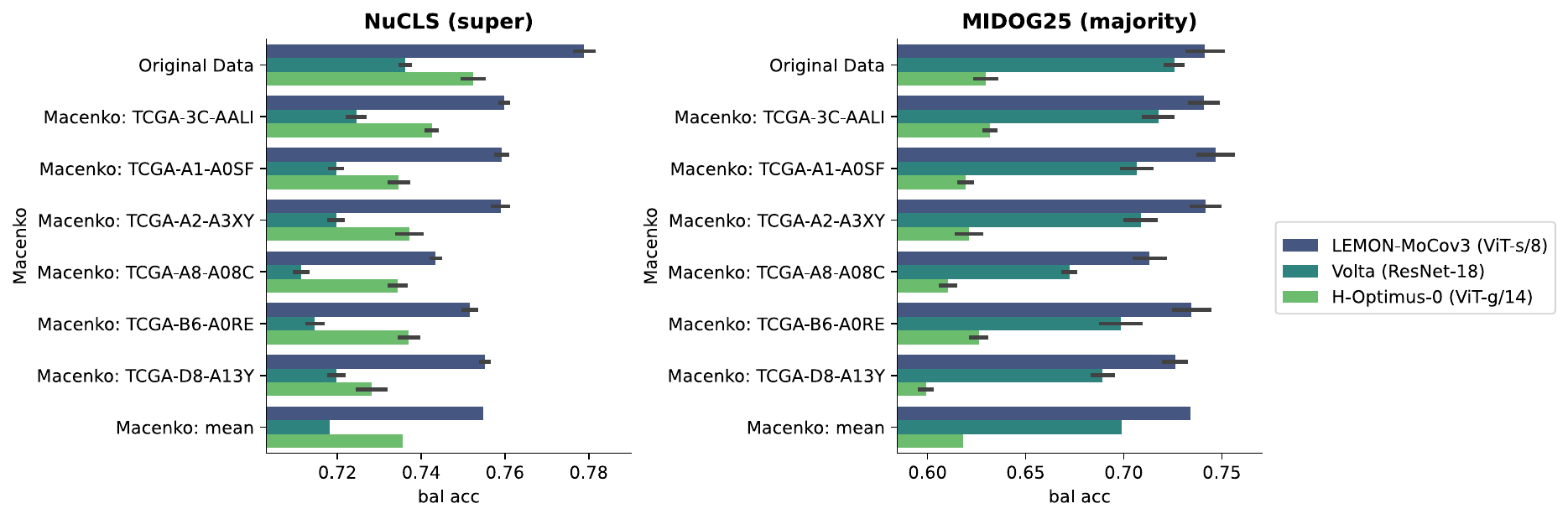}
    \caption{Benchmark of performances under different Macenko staining augmentations. Balanced accuracy of three models (H-optimus-0, LEMON, and Volta) on the NuCLS (left) and MIDOG25 (right) datasets. For each dataset, results are shown for the original images, six Macenko-stained augmentation conditions, as well as the mean across these six augmentations.}
    \label{supfig:macenko-perf}
\end{figure}

We further provide detailed performance comparisons across staining conditions for NuCLS and MIDOG25 (see \autoref{supfig:macenko-perf}). These results confirm that LEMON-MoCov3 maintains consistent downstream classification performance across stain perturbations, outperforming Volta and H-Optimus-0.

\section{Representation Analysis}\label{app-representationanlysis}

\begin{figure}
    \centering
    \includegraphics[width=0.75\textwidth]{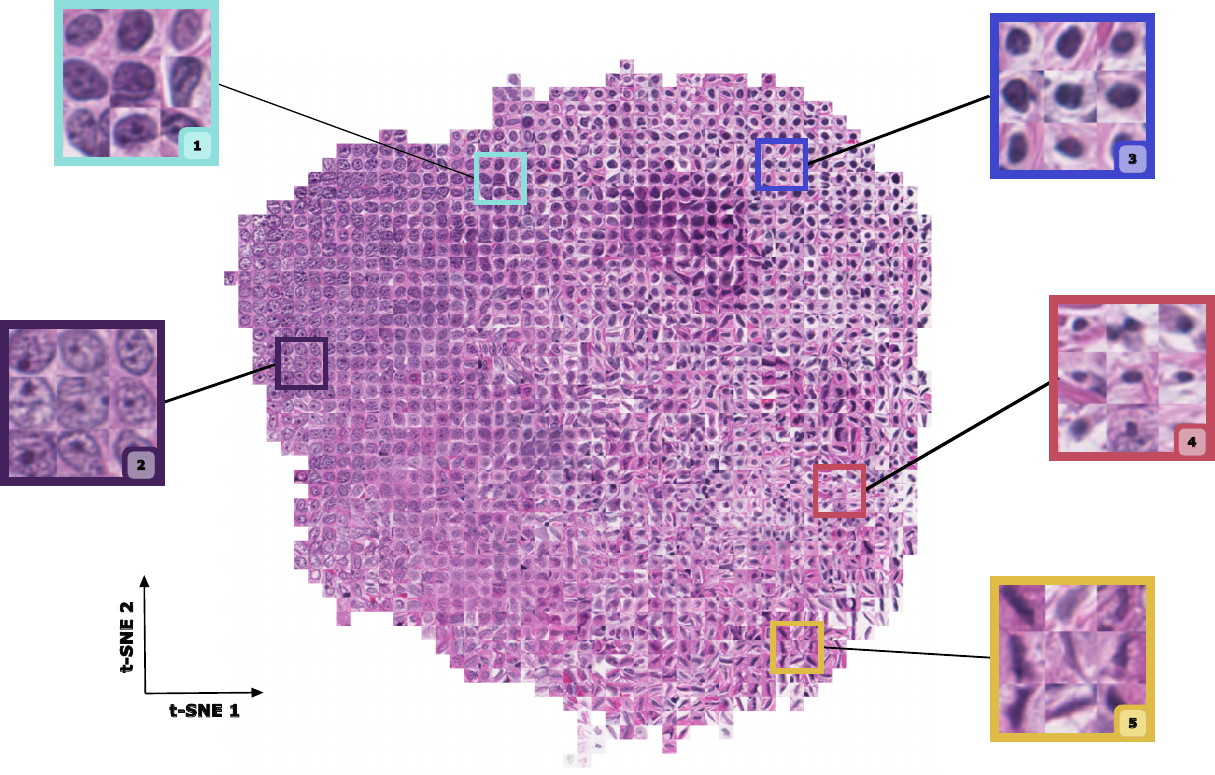}
    \caption{\textbf{Morphological map of learned cell representations.} A \ac{tsne} projection of 100,000 cells randomly sampled from a breast cancer whole-slide image (TCGA) is shown. A regular grid of cell images visualizes the embedding space, revealing a smooth continuum of latent morphologies. Five representative regions were annotated by an expert pathologist, corresponding to distinct cellular phenotypes: (1) epithelial cells, (2) neoplastic epithelial cells with large, irregular nuclei, (3) small lymphocytes, (4) plasma cells, and (5) fibroblasts.}
    \label{fig:tsne}
\vskip -0.3in
\end{figure}

We further examined the learned cell embeddings on a TCGA Whole Slide Image (WSI) to assess whether they capture meaningful biological structure. For this, we projected the embeddings into two dimensions using \ac{tsne} (\autoref{fig:tsne}). 

The spatial organization of the latent space revealed clear biological structure. Along the first dimension, cells located on the far left (1) were predominantly epithelial. Within this region, we identified a subpopulation in the central left area (2) characterized by large, rounded nuclei with irregular borders and prominent nucleoli, suggestive of neoplastic epithelial cells. In the upper right corner (3), we observed clusters of cells with small, round nuclei, consistent with lymphocytes. An adjacent cluster (4) likely corresponded  to plasma cells, with eccentric nuclei. Finally, the lower right region (5) contained elongated, spindle-shaped nuclei, suggestive of fibroblasts or myofibroblasts. 

These observations highlight that the model captures not only broad cell-type distinctions but also finer-grained morphological and pathological variations directly from the learned representations.

\section{Additional classification metrics for benchmarks}\label{app-additional_metrics}

\begin{table}[!hb]
\caption{Evaluation on the NuCLS dataset - task super}
\label{tab:nucls_full}
\centering
\resizebox{\textwidth}{!}{%
\begin{tabular}{llllll}
\toprule
Target & \multicolumn{5}{c}{NuCLS (super)} \\
Metric & auc & aupr & bal acc & f1 (macro) & f1 (weighted) \\
Embeddings &  &  &  &  &  \\
\midrule
\textbf{Imagenet (ResNet-18)} & 0.876 (0.003) & 0.757 (0.005) & 0.693 (0.005) & 0.698 (0.006) & 0.74 (0.005) \\
\textbf{Imagenet (ResNet-50)} & 0.891 (0.004) & 0.777 (0.006) & 0.711 (0.005) & 0.717 (0.005) & 0.756 (0.004) \\
\midrule
\textbf{DINOv2 (ViT-s/14)} & 0.905 (0.001) & 0.794 (0.002) & 0.732 (0.003) & 0.738 (0.004) & 0.775 (0.004) \\
\textbf{DINOv2 (ViT-b/14)} & 0.903 (0.003) & 0.792 (0.006) & 0.732 (0.003) & 0.738 (0.004) & 0.775 (0.003) \\
\textbf{DINOv2 (ViT-l/14)} & 0.897 (0.003) & 0.775 (0.005) & 0.721 (0.004) & 0.726 (0.005) & 0.767 (0.003) \\
\textbf{DINOv2 (ViT-g/14)} & 0.905 (0.004) & 0.786 (0.007) & 0.733 (0.008) & 0.739 (0.008) & 0.777 (0.007) \\
\midrule
\textbf{UNI (ViT-l/16)} & 0.905 (0.004) & 0.789 (0.006) & 0.733 (0.007) & 0.738 (0.007) & 0.777 (0.006) \\
\textbf{Virchow2 (ViT-h/14)} & 0.915 (0.004) & 0.807 (0.007) & 0.753 (0.008) & 0.759 (0.008) & 0.794 (0.007) \\
\textbf{H-Optimus-0 (ViT-g/14)} & 0.917 (0.002) & 0.812 (0.003) & 0.752 (0.006) & 0.757 (0.007) & 0.794 (0.006) \\
\midrule
\textbf{Volta (ResNet-18)} & 0.905 (0.003) & 0.795 (0.002) & 0.736 (0.003) & 0.74 (0.003) & 0.776 (0.002) \\
\textbf{LEMON-MoCov3 (ViT-s/8)} & \textbf{0.932 (0.003)} & \textbf{0.831 (0.004)} & \textbf{0.779 (0.005)} & \textbf{0.785 (0.005)} & \textbf{0.816 (0.005)} \\
\bottomrule
\end{tabular}

}
\end{table}

\begin{table}[!hb]
\caption{Evaluation on the MIDOG25 dataset}
\label{tab:midog25_full}
\centering
\resizebox{\textwidth}{!}{%
\begin{tabular}{llllll}
\toprule
Target & \multicolumn{5}{c}{MIDOG25 (majority)} \\
Metric & auc & aupr & bal acc & f1 (macro) & f1 (weighted) \\
Embeddings &  &  &  &  &  \\
\midrule
\textbf{Imagenet (ResNet-18)} & 0.781 (0.01) & 0.948 (0.001) & 0.577 (0.008) & 0.597 (0.012) & 0.829 (0.004) \\
\textbf{Imagenet (ResNet-50)} & 0.805 (0.014) & 0.955 (0.003) & 0.603 (0.012) & 0.632 (0.016) & 0.84 (0.006) \\ 
\midrule
\textbf{DINOv2 (ViT-s/14)} & 0.832 (0.011) & 0.961 (0.003) & 0.643 (0.004) & 0.677 (0.004) & 0.856 (0.002) \\
\textbf{DINOv2 (ViT-b/14)} & 0.842 (0.007) & 0.964 (0.001) & 0.653 (0.013) & 0.688 (0.013) & 0.86 (0.004) \\
\textbf{DINOv2 (ViT-l/14)} & 0.833 (0.012) & 0.961 (0.003) & 0.647 (0.01) & 0.682 (0.009) & 0.858 (0.003) \\
\textbf{DINOv2 (ViT-g/14)} & 0.844 (0.01) & 0.964 (0.003) & 0.654 (0.01) & 0.69 (0.009) & 0.861 (0.003) \\
\midrule
\textbf{UNI (ViT-l/16)} & 0.804 (0.012) & 0.953 (0.005) & 0.623 (0.008) & 0.656 (0.01) & 0.849 (0.005) \\
\textbf{Virchow2 (ViT-h/14)} & 0.799 (0.013) & 0.95 (0.006) & 0.618 (0.016) & 0.65 (0.02) & 0.848 (0.007) \\
\textbf{H-Optimus-0 (ViT-g/14)} & 0.812 (0.011) & 0.954 (0.003) & 0.631 (0.012) & 0.665 (0.015) & 0.852 (0.006) \\
\midrule
\textbf{Volta (ResNet-18)} & 0.849 (0.014) & 0.963 (0.005) & 0.726 (0.01) & 0.742 (0.006) & 0.876 (0.004) \\
\textbf{LEMON-MoCov3 (ViT-s/8)} & \textbf{0.911 (0.01)} & \textbf{0.981 (0.003)} & \textbf{0.741 (0.02)} & \textbf{0.778 (0.019)} & \textbf{0.897 (0.008)} \\
\bottomrule
\end{tabular}

}
\end{table}

\begin{table}[!hb]
\caption{Evaluation on the PanNuke dataset}
\label{tab:pannuke_full}
\centering
\resizebox{\textwidth}{!}{%
\begin{tabular}{llllll}
\toprule
Target & \multicolumn{5}{c}{PanNuke (classification)} \\
Metric & auc & aupr & bal acc & f1 (macro) & f1 (weighted) \\
Embeddings &  &  &  &  &  \\
\midrule
\textbf{Imagenet (ResNet-18)} & 0.873 (0.002) & 0.589 (0.007) & 0.56 (0.004) & 0.583 (0.005) & 0.643 (0.003) \\
\textbf{Imagenet (ResNet-50)} & 0.885 (0.002) & 0.622 (0.005) & 0.587 (0.005) & 0.608 (0.005) & 0.663 (0.003) \\
\midrule
\textbf{DINOv2 (ViT-s/14)} & 0.904 (0.002) & 0.661 (0.006) & 0.635 (0.009) & 0.653 (0.008) & 0.7 (0.003) \\
\textbf{DINOv2 (ViT-b/14)} & 0.901 (0.001) & 0.655 (0.005) & 0.627 (0.01) & 0.646 (0.008) & 0.691 (0.003) \\
\textbf{DINOv2 (ViT-l/14)} & 0.899 (0.003) & 0.644 (0.009) & 0.619 (0.008) & 0.639 (0.007) & 0.691 (0.003) \\
\textbf{DINOv2 (ViT-g/14)} & 0.904 (0.002) & 0.652 (0.006) & 0.631 (0.007) & 0.651 (0.006) & 0.701 (0.002) \\
\midrule
\textbf{UNI (ViT-l/16)} & 0.91 (0.002) & 0.689 (0.005) & 0.643 (0.008) & 0.663 (0.007) & 0.704 (0.003) \\
\textbf{Virchow2 (ViT-h/14)} & 0.914 (0.002) & 0.685 (0.007) & 0.645 (0.009) & \textbf{0.666 (0.008)} & 0.708 (0.004) \\
\textbf{H-Optimus-0 (ViT-g/14)} & \textbf{0.915 (0.002)} & \textbf{0.691 (0.005)} & \textbf{0.648 (0.007)} & 0.665 (0.006) & 0.71 (0.002) \\
\midrule
\textbf{Volta (ResNet-18)} & 0.876 (0.002) & 0.565 (0.002) & 0.571 (0.007) & 0.581 (0.007) & 0.669 (0.002) \\
\textbf{LEMON-MoCov3 (ViT-s/8)} & 0.912 (0.001) & 0.643 (0.005) & 0.641 (0.008) & 0.66 (0.007) & \textbf{0.712 (0.003)} \\
\bottomrule
\end{tabular}

}
\end{table}

\section{Use of Large Language Models (LLMs)}

We used LLMs to aid and polish writing.

\end{document}